%% file: sample-sigconf.tex
\documentclass[sigconf]{acmart}
\usepackage{glossaries}
\usepackage{adjustbox}

\usepackage{algorithm}
\usepackage{algorithmicx}
\usepackage{algpseudo code}
\usepackage{subfigure}
\usepackage{graphicx}
\usepackage{syntax}
\usepackage[font=small]{caption}

\AtBeginDocument{%
  \providecommand\BibTeX{{%
    \normalfont B\kern-0.5em{\scshape i\kern-0.25em b}\kern-0.8em\TeX}}}

\copyrightyear{2022}
\acmYear{2022}
\setcopyright{acmlicensed}\acmConference[GECCO '22]{Genetic and Evolutionary Computation Conference}{July 9--13, 2022}{Boston, MA, USA}
\acmBooktitle{Genetic and Evolutionary Computation Conference (GECCO '22), July 9--13, 2022, Boston, MA, USA}
\acmPrice{15.00}
\acmDOI{10.1145/3512290.3528833}
\acmISBN{978-1-4503-9237-2/22/07}

\begin{document}

\input{acronyms}
\title{Co-evolutionary Probabilistic Structured Grammatical Evolution}

\author{Jessica Mégane}
\email{jessicac@dei.uc.pt}
\orcid{0000-0001-6697-5423}

\affiliation{%
  \institution{University of Coimbra}
  \city{Coimbra}
  \streetaddress{Polo II - Pinhal de Marrocos}
  \country{Portugal}
  \postcode{3030}
}

\author{Nuno Lourenço}
\orcid{0000-0002-2154-0642}
\email{naml@dei.uc.pt}
\affiliation{%
  \institution{University of Coimbra}
  \city{Coimbra}
  \streetaddress{Polo II - Pinhal de Marrocos}
  \country{Portugal}
  \postcode{3030}
}

\author{Penousal Machado}
\orcid{0000-0002-6308-6484}
\email{machado@dei.uc.pt}
\affiliation{%
  \institution{University of Coimbra}
  \city{Coimbra}
  \streetaddress{Polo II - Pinhal de Marrocos}
  \country{Portugal}
  \postcode{3030}
}



\input{text/abstract}




\maketitle

\input{text/intro}

\input{text/ge}

\input{text/co}

\input{text/performance}

\input{text/results}

\input{text/conc}
\begin{acks}
This work is funded by national funds through the FCT - Foundation for Science and Technology, I.P., within the scope of the project CISUC - UID/CEC/00326/2020 and within the scope of the project A4A: Audiology for All (CENTRO-01-0247-FEDER-047083) financed by the Operational Program for Competitiveness and Internationalisation of PORTUGAL 2020.
\end{acks}

\bibliographystyle{ACM-Reference-Format}
\bibliography{sample-base}


\end{document}

%% file: acronyms.tex
\newacronym{ge}{GE}{Grammatical Evolution}
\newacronym{gp}{GP}{Genetic Programming}
\newacronym{pge}{PGE}{Probabilistic Grammatical Evolution}
\newacronym{sge}{SGE}{Structured Grammatical Evolution}
\newacronym{dsge}{DSGE}{Dynamic Structured Grammatical Evolution}
\newacronym{ea}{EA}{Evolutionary Algorithm}
\newacronym{cfg}{CFG}{Context-Free Grammar}
\newacronym{pcfg}{PCFG}{Probabilistic Context-Free Grammar}
\newacronym{eda}{EDA}{Estimation Distribution Algorithm}
\newacronym{copge}{Co-PGE}{Co-evolutionary Probabilistic Grammatical Evolution}
\newacronym{pmbge}{PMBGE}{Probabilistic Model Building Grammatical Evolution}
\newacronym{cdt}{CDT}{Conditional Dependency Tree}
\newacronym{nt}{NT}{Non-terminal}
\newacronym{pige}{$\pi$GE}{Position Independent Grammatical Evolution}
\newacronym{cfggp}{CFG-GP}{Context-Free Grammar Genetic Programming}
\newacronym{rrse}{RRSE}{Root Relative Squared Error}
\newacronym{ec}{EC}{Evolutionary Computation}
\newacronym{adf}{ADF}{Automatically Defined Functions}
\newacronym{psge}{PSGE}{Probabilistic Structured Grammatical Evolution}
\newacronym{tag}{TAG}{Tree-Adjunct Grammar}
\newacronym{tage}{TAGE}{Tree-Adjunct Grammatical Evolution}
\newacronym{pigrow}{PI Grow}{Position Independent Grow}
\newacronym{ptc2}{PTC2}{Probabilistic Tree Creation 2}
\newacronym{mi}{MI}{Mutation Innovation}
\newacronym{ci}{CI}{Crossover Innovation}
\newacronym{ai}{AI}{Artificial Intelligence}
\newacronym{copsge}{Co-PSGE}{Co-evolutionary Probabilistic Structured Grammatical Evolution}
\newacronym{mgga}{mGGA}{meta-Grammar Genetic Algorithm}
\newacronym{ge2}{(GE)$^2$}{Grammatical Evolution by Grammatical Evolution}
\newacronym{autoge}{AutoGE}{Automatic Grammatical Evolution}

%% file: text/abstract.tex
\begin{abstract}

This work proposes an extension to \gls{sge} called \gls{copsge}.
In \gls{copsge} each individual in the population is composed by a grammar and a genotype, which is a list of dynamic lists, each corresponding to a non-terminal of the grammar containing real numbers that correspond to the probability of choosing a derivation rule. Each individual uses its own grammar to map the genotype into a program. During the evolutionary process, both the grammar and the genotype are subject to variation operators.

The performance of the proposed approach is compared to 3 different methods, namely, \gls{ge}, \gls{pge}, and \gls{sge} on four different benchmark problems. The results show the effectiveness of the approach since \gls{copsge} is able to outperform all the methods with statistically significant differences in the majority of the problems.





\end{abstract}

%% file: text/intro.tex
\glsresetall
\section{Introduction}
\glspl{ea} is the name given to a set of stochastic search procedures that are loosely inspired by the principles of natural selection and genetics. These methods iteratively improve a set of candidate solutions, usually referred to as the population, guided by a fitness function. The improvement is obtained by selecting the most promising solutions (taking into account the objective function), and applying some stochastic variations using operators similar to mutations and recombinations that take place in biological reproduction.
Individuals with higher fitness are more likely to survive and reproduce. The application of these elements is repeated for several generations and it is expected that, over time, the quality of individuals improves.

Grammars are used by many \glspl{ea} to define the search space and impose limitations on the solutions. The same syntax can be represented by an infinite amount of grammars, so a good grammar design may impact on the effectiveness of the optimization by converging to better solutions faster.
\gls{cfggp} \cite{whigham} and \gls{ge} \cite{ONeill2003,Ryan1998,handbookge} are the most popular grammar-based \gls{gp} algorithms.
Both these methods use grammars to introduce syntactic constraints, but the way individuals are represented differs. Individuals in \gls{cfggp} are derivation trees that start with the axiom of the grammar. In \gls{ge}, individuals have a genotype, which is a vector of integers, that with the help of the \gls{cfg}, is used to expand a derivation rule to form the solution (i.e., phenotype).

\gls{ge} has grown in popularity over time, although it does have some drawbacks, such as low locality and high redundancy \cite{Rothlauf2003,Rothlauf2006}.
When several genotypes map to the same phenotype, a representation is said to have high redundancy. Locality has to do with how variations in the genotype cause phenotypic alterations. The best scenario, i.e. high locality, would be such that a small change in the genotype would result in a small change in the phenotype.
Because of these two issues, exploitation is often changed by exploration, behaving similarly as random search in some problems \cite{Whigham2015}. However, some \gls{ge} alternatives have developed in the literature that have improved results by introducing different mechanisms for representing individuals \cite{ONeill2004,Kim2015,KIM2016,Lourenco2016,Loureno2017,Ryan2002,Megane2021}, population initialization \cite{Nicolau2017,ryanazad2003,Fagan2016,Luke2000,Murphy2012} and grammar design \cite{Nicolau2018,Harper2010,nicolau2004,ONeillRyan2004,ONeill2005,ali2021autoge}.


In the literature it has been shown that adjusting the biases of the grammar productions based on individuals in the population can improve the results \cite{Megane2021,KIM2016}, as can co-evolving the grammar with the genetic code \cite{ONeillRyan2004}.
Due to the impact that grammars can have on the quality and variability of the generated solutions, the goal of this work was to create a method that evolves the probabilities of the grammar, also introducing some randomness to prevent it from converging to local optima solutions.

In this work we propose \gls{copsge}, an extension of \gls{sge} \cite{Lourenco2016,Loureno2018}, in which each individual is composed of its genotype and a grammar. \gls{sge} is a method that besides presenting better performance than \gls{ge}, it also overcomes its issues, presenting high locality and low redundancy \cite{Loureno2016}. 
The genotype of the \gls{copsge}  individuals is a set of lists, one for each non-terminal of the grammar, and each element of the list corresponds to the probability of choosing a production rule of the respective non-terminal.
The grammar of each individual can be subject to variation operators, and at the end of each generation it is used to update its phenotype.

The proposed method was compared with standard \gls{ge} \cite{ONeill2003,Ryan1998,handbookge}, \gls{pge} \cite{Megane2021} and \gls{sge} \cite{Lourenco2016,Loureno2018} on four different benchmark problems \cite{McDermott2012}. Statistical tests show that \gls{copsge} is better than \gls{ge}, outperforming it with significant differences on all problems. \gls{pge} was outperformed in three of the four problems analyzed, while \gls{sge} was outperformed in two problems.

The remainder of this work is structured as follows: Section \ref{sec:ge} presents the background necessary to understand the work presented, introducing \gls{ge} as well as related work.
Section \ref{sec:copsge} present the proposed method, detailing the representation and mapping method used. Section \ref{sec:set} details the experimentation framework used and Section \ref{sec:res} the experimental results regarding performance. Section \ref{sec:conc} gathers the main conclusions and provides some insights regarding future work.

%% file: text/ge.tex
\section{Grammatical Evolution}
\label{sec:ge}
\gls{ge} \cite{ONeill2003,Ryan1998,handbookge} is a grammar-based \gls{ea} in which individuals are represented as a list of integers. The elements of the list (i.e., codons) are mapped creating the phenotype of the individual (i.e., solution to the problem), using production rules defined by a \gls{cfg}.
A grammar is a tuple $G = (NT,T,S,P)$ where $NT$ and $T$ represent the non-empty set of \textit{Non-Terminal (NT)} and \textit{Terminal (T)} symbols, $S$ is an element of $NT$ in which the derivation sequences start, called the axiom, and $P$ is the set of production rules. The rules in $P$ are in the form $A ::= \alpha$, with $A \in NT$ and $\alpha \in (NT \cup T)^*$. The $NT$ and $T$ sets are disjoint. Each grammar defines a language $L(G)= \{ w:\, S\overset{*} {\Rightarrow} w,\, w \in T^*\}$, that is the set of all sequences of terminal symbols that can be derived from the axiom. The symbol $*$ represents the unary operator Kleene star.

The genotype of individuals is a list of integer values (i.e., codons) randomly generated in the interval $[0, 255]$.
The genotype-phenotype mapping starts with the axiom of the grammar, and the expansion is always made from the leftmost non-terminal. Each codon is mapped into a production rule by applying the modulo operator ($mod$) between the codon and the number of expansion options of the non-terminal to expand.

\begin{figure}[h!]
		\scalebox{0.7}{\fbox{%
				\parbox{0.1\textwidth}{%
					\begin{align*}
						{<}\text{expr}{>} ::=  & \, {<}\text{expr}{>}{<}\text{op}{>}{<}\text{expr}{>} \\
						& | \, {<}\text{var}{>} \\
						{<}\text{op}{>} ::= & \,+ \, \\
						& | \, - \\
						& | \, * \\
						& | \, / \\
						{<}\text{var}{>} ::= &  \, \text{x} \, \\ 
						& | \, \text{y} \,  \\
						& | \, \text{1.0} \,
		\end{align*}}}}
		\caption{\label{gemapgram}Example of grammar.}
	\end{figure}

\begin{figure}[h!]
		\scalebox{0.7}{\fbox{%
				\parbox{0.1\textwidth}{%
					\begin{align*}
						\textbf{Gen} & \textbf{otype}\\
						{[}\text{54, 7, 83, 237, 71, 123, } & \text{67, 142, 25, 195, 202, 153}{]}\\
						{<}\text{expr}{>} \rightarrow  & \, {<}\text{expr}{>}{<}\text{op}{>}{<}\text{expr}{>} & 54 mod 2 = 0 \\
						{<}\text{expr}{>}{<}\text{op}{>}{<}\text{expr}{>} \rightarrow & \, {<}\text{var}{>}{<}\text{op}{>}{<}\text{expr}{>} & 7 mod 2 = 1 \\
						{<}\text{var}{>}{<}\text{op}{>}{<}\text{expr}{>} \rightarrow & \, \text{1.0} \, {<}\text{op}{>}{<}\text{expr}{>} & 83 mod 3 = 2\\
						\text{1.0}{<}\text{op}{>}{<}\text{expr}{>} \rightarrow & \, \text{1.0} \, \text{-} \, {<}\text{expr}{>} & 237 mod 4 = 1\\
						\text{1.0} \, \text{-} \, {<}\text{expr}{>} \rightarrow & \, \text{1.0} \, \text{-} \, {<}\text{var}{>} & 71 mod 2 = 1 \\
						\text{1.0} \, \text{-} \, {<}\text{var}{>} \rightarrow & \, \text{1.0} \, \text{-} \, \text{x} & 123 mod 3 = 0 \\
						\textbf{Phenotype: } & 1.0 - x
		\end{align*}}}}
	\caption{\label{fig:gemapping}Example of the genotype-phenotype mapping of GE.}
\end{figure}

An example of this process is shown in Fig. \ref{fig:gemapping}, using the example of grammar presented in Fig. \ref{gemapgram}.
The mapping begins with $<$expr$>$, the grammar's axiom, which has two expansion alternatives, and the genotype's first unused value, $54$. Applying the mapping function, $54 mod(2) = 0$, we obtain the index of the rule to be selected, $<$expr$><$op$><$expr$>$.
This process is performed until there are no more non-terminal symbols to expand or there are no more integers to read from the genotype.
In this last case and if we still have non-terminals to expand, a wrapping mechanism can be used, where the genotype will be re-used until it generates a valid individual or the predefined number of wraps is over. If after all the wraps we still have not mapped all the non-terminals, the mapping process stops, and the individual will be considered invalid. The phenotype of each individual is evaluated with the fitness function and then the population goes through the selection mechanisms.

\subsection{Related Work}
\label{sec:related}

\gls{ge} is one of the most used \gls{gp} variants, and over the years it has been subject of several improvements to address some of of its main criticisms, namely the high redundancy and low locality. Both properties are related to the impact of genetic operators on an individual's phenotype, with low locality referring to small genotype changes that make big changes in the phenotype, while high redundancy means that there are many genotypes corresponding to the same phenotype.
The majority of these proposed solutions include changes to grammars \cite{Harper2010,nicolau2004,Nicolau2018,ONeillRyan2004,ONeill2005,ali2021towards}, individual representation \cite{Kim2015,KIM2016,Loureno2018,Megane2021,ONeill2004,Ryan2002}, or population initialization \cite{Fagan2016,Luke2000,Murphy2012,Nicolau2017,ryanazad2003}.

\gls{sge} \cite{Loureno2018} is a recent proposal that tackles the locality and redundancy issues of \gls{ge} \cite{Lourenco2016}, at the same time achieving better performance results \cite{Loureno2017}. 
In \gls{sge} the genotype is a set of dynamic lists of ordered integers, with one list for each non-terminal of the grammar. Each value in the list represents which production rule to choose from the respective non-terminal.
In \cite{Loureno2017} different grammar-based \gls{gp} approaches were compared, and the authors showed that \gls{sge} achieved a good performance when compared with several grammar-based \gls{gp} representations.

\gls{pige} \cite{ONeill2004} is a method that introduces a different representation and mapping mechanism, in which the order of expansion of the non-terminals is determined by the genotype of the individual, removing the positional dependency that exists in \gls{ge}. The genotype of the individuals is composed of two values (\textit{nont}, \textit{rule}), where \textit{nont} used to select the next non-terminal to be expanded, and \textit{rule} selects which rule to derive from that non-terminal. This method proved to be better than \gls{ge} on several problems, showing statistical differences \cite{Fagan2010}.

Chorus \cite{Ryan2002} is another method in which there is positional independence, with each gene specifically encoding one production of the grammar. However, this approach has not been shown to be better than the \gls{ge} standard.

The design of the grammars is another aspect that has had some attraction for researchers, since they define the search space, and so the choice of grammar can influence the speed of convergence to the best solution \cite{Nicolau2018}. 
Some studies have been conducted to analyze the performance of \gls{ge} with different types of grammars, such as the use of recursively balanced grammars \cite{Harper2010,Nicolau2018} and the reduction of non-terminal symbols \cite{nicolau2004,Nicolau2018}.

Harper et al. \cite{Harper2010} showed that the grammar chosen at the beginning of the evolutionary process can have a large impact on the solutions, such as generating many invalid individuals when using recursive grammars. It has also been shown that there is more variety in the size of solutions when using a balanced grammar.

Nicolau et al. \cite{Nicolau2018} tested \gls{ge} using different types of grammars, which included, balanced grammars, grammars with corrected biases, and grammars with unlinked productions.
The tests using recursively balanced grammars, in which for every recursive production there is a non-recursive one, showed improved results over the original grammar. However it resulted in a larger number of individuals consisting of a non-terminal symbol.
Nicolau \cite{nicolau2004} proposed a method to reduce the number of non-terminal symbols, replacing them by their productions, which despite showing a slight increase in performance, it has the disadvantage of generating very complex grammars that are difficult to read.

Another line of study has been the evolution of grammar during the evolutionary process \cite{ONeillRyan2004,Megane2021,Kim2015,KIM2016,ali2021autoge}.

\gls{ge2} \cite{ONeillRyan2004} is an approach in which there is co-evolution of grammar and genetic code. The method uses two distinct grammars, the universal grammar and the solution grammar. The universal grammar dictates the structure of the solution grammar, that is used to map the individuals. This method has shown to be effective in evolving biases towards some non-terminal symbols. Later, was implemented into a new algorithm, \gls{mgga} \cite{ONeill2005}, which obtained performance improvements. 

\gls{autoge} \cite{ali2021autoge} is a tool designed to help define the structure of the grammar and fitness function to be used by \gls{ge}. This tool introduces a Production Rule Pruning (PRP) algorithm, which assigns ranks to productions to detect the least valuable ones. This strategy was tested on real-world symbolic regression problems \cite{ali2021towards}, and obtained significant performance improvements in 3 of the 10 problems analyzed. Although in all problems there were improvements in genotype size, these were significant in 7 of the 10 problems.

\gls{pge} \cite{Megane2021} is a recent variant of \gls{ge}, in which a \gls{pcfg} is used to map the individuals and the genotype is a list of real numbers. A \gls{pcfg} is a quintuple $PG = (NT,T,S,P,Probs)$ where NT and T represent the non-empty set of \textit{Non-Terminal (NT)} and \textit{Terminal (T)} symbols, $S$ is an element of $NT$ called the axiom, $P$ is the set of production rules, and $Probs$ is a set of probabilities associated with each production rule. The mapping is done from the leftmost non-terminal, and for each non-terminal to be expanded, the rule whose probability interval includes the codon is selected. At the end of each generation, the \gls{pcfg} probabilities are updated based on the expansion rules used to create the best individual of the current generation alternating with the best individual overall. \gls{pge} proved to be better than \gls{ge} with statistical differences in the two problems analysed.

Kim et al. \cite{Kim2015} have proposed a \gls{pmbge} in which the mapping uses a \gls{pcfg} to the probabilistic technique \gls{eda}, which also replaces the mutation and crossover operators. This technique at each generation generates a new population from the new grammar, whose probabilities are generated based on the frequency of the rules expanded by the best individuals. The proposed approach had a similar performance when compared with \gls{ge}. 
Later, Kim et al. \cite{KIM2016} adapted \gls{cdt} to the mechanism of updating the grammar, in which dependencies between production rules are considered. This method outperformed \gls{ge} with statistical differences in two of the four problems analysed.
The major distinction between these two methods and \gls{pge} \cite{Megane2021} is in the method of adjusting the probabilities of the grammar over the evolutionary process and that \gls{pge} considers the genetic operators while the two other methods uses \gls{eda}.

The main difference between these two methods and \gls{pge} \cite{Megane2021} is in the mechanism of altering the grammar's probability over the evolutionary process, as well as the fact that \gls{pge} considers genetic operators, whilst the other two methods employ \gls{eda}.

%% file: text/co.tex
\section{Co-Evolutionary Probabilistic Structured Grammatical Evolution}
\label{sec:copsge}
In this paper we propose \gls{copsge}\footnote{The implementation of Co-PSGE is available at: \url{https://github.com/jessicamegane/co-psge}}, an extension to \gls{sge} that adapts its representation and mapping mechanism to use a \gls{pcfg}, which, like the individuals in the population, can be subject to variation operators.

Each individual in the population is composed of a grammar and its genotype, which is a set of dynamical lists, each list being associated with a non-terminal of the grammar. Each list contains an ordered sequence of real numbers, bounded to the interval [0, 1], that corresponds to the probability of selecting a derivation rule.
Each individual uses its own grammar to map the genotype into a program.
During the evolutionary process both the grammar and the genotype are subject to variation operators.
At the end of each generation, the phenotype is updated using the new genotype and its updated grammar.

At initialisation, a genotype is generated for each individual as well as a \gls{pcfg}. The grammar assigned to the individual at initialization has the same probability for each production rule of each non-terminal.
Individuals are initialized recursively, with a codon generated randomly and added to the genotype list of the non-terminal that is being expanded at each iteration (Alg. \ref{psgeindividual}, lines 2-3). The genotype (which starts empty for each non-terminal), the non-terminal symbol to expand (in the first iteration, the axiom of the grammar), the current depth (which starts at 0), the maximum depth limit and the individual's grammar are all given as parameters to the algorithm. The function simulates the mapping process, which will be described in detail further on (Alg. \ref{psgeindividual}, line 4) to determine which non-terminals should be expanded next. When all non-terminals symbols have been expanded (i.e., only terminals remain) the algorithm terminates with a valid individual.

\begin{algorithm}[h!]
  \caption{Random Candidate Solution Generation of Co-PSGE}
	\label{psgeindividual}
	\begin{algorithmic}[1]
		\Procedure{createIndividual}{genotype, symb, current\_depth, max\_depth, pcfg}
		\State codon = $random$(0,1)
		\State genotype[symb].append(codon)
		\State selected\_rule = $generate\_expansion$(symb, codon, pcfg, current\_depth, max\_depth)
		\State expansion\_symbols = pcfg[symb][selected\_rule]
		\For{sym \textbf{in} expansion\_symbols}
		\If{\textbf{not} $is\_terminal$(sym)}
		\State $createIndividual$(genotype, sym, current\_depth + 1, max\_depth, pcfg)
		\EndIf
		\EndFor
		\EndProcedure
	\end{algorithmic}
\end{algorithm}


The genotype-phenotype mapping process is presented in Alg. \ref{psgemapping}. The algorithm receives as arguments the genotype, a counter called $positions\_to\_map$ (which is used to store the genotype position of each non-terminal list at the current iteration, initialized with 0 for each non-terminal), the symbol to expand (starts in the axiom), the current depth, the maximum depth limit and the grammar.
If, during mapping, more codons are needed to create a valid individual, they will be randomly generated and added to the genotype (Alg. \ref{psgemapping}, lines 3-6). The dynamic genotype is one of the advantages of the representation proposed by \gls{sge} \cite{Loureno2018}. With the depth limit, it is possible to add productions whenever necessary, without the risk of bloat (a considerable growth in the size of the solutions \cite{Eiben2015}), always creating valid individuals.

\begin{algorithm}[h!]
	\caption{Genotype-Phenotype Mapping Function of Co-PSGE}
	\label{psgemapping}
	\begin{algorithmic}[1]
		\Procedure{mapping}{genotype, positions\_to\_map, symb, depth, max\_depth, pcfg}
		\State phenotype = ""
		\If{positions\_to\_map[symb] $>=$ $len$(genotype[symb])}
		\State codon = $random$(0,1)
		\State genotype[symb].append(codon)
		\EndIf
		\State codon = genotype[symb][positions\_to\_map[symb]]
		\State selected\_rule = $generate\_expansion$(symb, codon, pcfg, current\_depth, max\_depth)
		
		\State expansion = pcfg[symb][selected\_rule]
		\State positions\_to\_map[symb] += 1
		\For{sym \textbf{in} expansion}
		\If{$is\_terminal$(sym)}
		\State phenotype += sym
		\Else
		\State phenotype += $mapping$(genotype, positions\_to\_map, sym, depth + 1, max\_depth, pcfg)
		\EndIf
		\EndFor
		\State \textbf{return} phenotype
		\EndProcedure
	\end{algorithmic}
\end{algorithm}

The process of choosing a derivation rule from a codon of the genotype using a \gls{pcfg} is similar to that proposed by \gls{pge} \cite{Megane2021} (Alg. \ref{psgerule}, lines 15-21), except that there is a distinction when the maximum depth limit is exceeded, in which only non-recursive productions can be chosen (Alg. \ref{psgerule}, lines 4-13). The function receives as parameters the non-terminal symbol to be expanded, the codon, the grammar, the current depth and the maximum depth established.

When the defined maximum tree depth limit is exceeded (Alg. \ref{psgerule}, lines 4-13), the algorithm considers only non-recursive rules, and adjusts the probabilities of each of them, so that the sum is 1. To accomplish this, we first sum the value of the current probabilities of the non-recursive rules, which is used to perform the adjustment. Using the new probabilities, the production rule is chosen with the normal procedure: It is verified whether the codon belongs to the probability range of each production rule of the non-terminal to be expanded and when this condition is verified, the rule is chosen.

\begin{algorithm}[h!]
	\caption{Co-PSGE function to select an expansion rule}
	\label{psgerule}
	\begin{algorithmic}[1]
		\Procedure{generate\_expansion}{symb, codon, pcfg, depth, max\_depth}
		\State cum\_prob = 0.0
		\If{depth $>=$ max\_depth}
		\State nr\_prods = $get\_non\_recursive\_prods$(pcfg[symb])
		\State total\_nr\_prods = $sum$(nr\_prods.$getProb$())
		\For{prod \textbf{in} non\_recursive\_prods}
		\State new\_prob = prod.$getProb$() / total\_nr\_prods
		\State cum\_prob = cum\_prob + new\_prob
		\If{codon $\leq$ cum\_prob}
		\State selected\_rule = prod 
		\State \textbf{break}
		\EndIf
		\EndFor
		\Else
		\For{prod \textbf{in} pcfg[symb]}
		\State cum\_prob = cum\_prob + prod.$getProb()$
		\If{codon $\leq$ cum\_prob}
		\State selected\_rule = prod
		\State \textbf{break}
		\EndIf
		\EndFor
		\EndIf
		\State \textbf{return} selected\_rule
		\EndProcedure
	\end{algorithmic}
\end{algorithm}


The mapping process in \gls{copsge} is illustrated in Fig. \ref{fig:mappingcopsge}, using the \gls{pcfg} shown on Fig. \ref{copsgemapgram}. 
The mapping begins with the grammar's axiom, $<$expr$>$ and the first entry in the genotype's respective list, $0.29$.
The probabilities of the rules of the non-terminal are compared to the value of the codon. In this scenario, we have two equally probable possibilities. The non-terminal will be expanded to $<$expr$><$op$><$exp$>$ because $0.29$ falls within the range of probabilities of the initial production.

The next non-terminal to expand, according to the leftmost derivation rule, is $<$expr$>$.
The process is repeated, this time with the codon 0.73, which corresponds to the second derivation rule, $<$var$>$.
This non-terminal is the next to be expanded, and it has three derivation rules. $0.41$ is the first codon available in the list of the non-terminal $<$var$>$. The codon falls within the range of the second rule, $y$, when we look at the probabilities of the derivation rules. The procedure is repeated until a valid individual is formed.

\begin{figure}[h!]
	\noindent\scalebox{0.70}{\fbox{%
			\parbox{0.1\textwidth}{%
				\begin{align*}
					{<}\text{expr}{>} ::=  & \, {<}\text{expr}{>}{<}\text{op}{>}{<}\text{expr}{>} & (0.5)\\
					& | \, {<}\text{var}{>} & (0.5)\\
					{<}\text{op}{>} ::= & \,+ \, & (0.25)\\
					& | \, - & (0.25)\\
					& | \, * & (0.25)\\
					& | \, / & (0.25)\\
					{<}\text{var}{>} ::= &  \, \text{x} & (0.33) \\ 
					& | \, \text{y} & (0.33)  \\
					& | \, \text{1.0} & (0.33)
	\end{align*}}}}
\caption{\label{copsgemapgram}PCFG example.}
\end{figure}

\begin{figure}[h!]
\centering
\scalebox{0.70}{\fbox{%
		\parbox{0.1\textwidth}{%
			\begin{align*}
				\textbf{Genot} & \textbf{ype}\\ 
				\framebox[2.8cm]{$<expr>$}\framebox[2cm]{$<op>$}&\framebox[2.5cm]{$<var>$} \\
				\framebox[2.8cm]{[0.29,0.73,0.52]}\framebox[2cm]{[0.86]}&\framebox[2.5cm]{[0.41, 0.15]} \\
				{<}\text{expr}{>} \rightarrow  & \, {<}\text{expr}{>}{<}\text{op}{>}{<}\text{expr}{>} & (0.29) \\
				{<}\text{expr}{>}{<}\text{op}{>}{<}\text{expr}{>} \rightarrow & \, {<}\text{var}{>}{<}\text{op}{>}{<}\text{expr}{>} & (0.73)\\
				{<}\text{var}{>}{<}\text{op}{>}{<}\text{expr}{>} \rightarrow & \, \text{y} \, {<}\text{op}{>}{<}\text{expr}{>} & (0.41)\\
				\text{y}{<}\text{op}{>}{<}\text{expr}{>} \rightarrow & \, \text{y} \, \text{/} \, {<}\text{expr}{>} & (0.86)\\
				\text{y} \, \text{/} \, {<}\text{expr}{>} \rightarrow & \, \text{y} \, \text{/} \, {<}\text{var}{>} & (0.52) \\
				\text{y} \, \text{/} \, {<}\text{var}{>} \rightarrow & \, \text{y} \, \text{/} \, \text{x} & (0.15) \\
				\textbf{Phenotype: } & y / x
\end{align*}}}}

	\caption{Example of the genotype-phenotype mapping of Co-PSGE with a PCFG.}
	\label{fig:mappingcopsge}
\end{figure}


At the end of each generation the \gls{pcfg} of each individual is itself subject to variation through mutations on its probabilities. 
This process is demonstrated in Alg. \ref{algocopge}. The algorithm takes as parameters the individual whose grammar is to be mutated, the probability of mutation occurring (between 0 and 1) and the float corresponding to the standard deviation to use in a normal destribution.
The mutation probability is used to check for each non-terminal production rule whether mutation should occur (Alg. \ref{algocopge}, line 5).
When a production is selected for mutation, a generated value is added to it with a Gaussian distribution of mean 0 and the standard deviation previously defined, keeping the new value in the range [0,1] (Alg. \ref{algocopge}, lines 6-8). Gaussian mutations have been widely used in the literature and have showed to be a good approach to make small changes in the search space \cite{Backevolutionaryprogramming,hinterdinggaussian,Beyer2004EvolutionS}.
In this method only one mutation can occur per non-terminal, so after a production is mutated, the probabilities of the remaining productions of the respective non-terminal are updated until their sum equals 1, and can only be changed again in the next generation (Alg. \ref{algocopge}, line 9).
The grammar of the parent with the best fitness is passed on to the offspring during crossover.
The individual's phenotype is updated at the end of each generation using the new grammar and genotype.

\begin{algorithm}
	\caption{Co-PSGE algorithm to mutate PCFG's probabilities.}
	\label{algocopge}
	\begin{algorithmic}[1]
		\Procedure{mutateGrammar}{individual, prob\_mutation, sd\_normal\_dist}
		\State $grammar = individual.getGrammar()$
		
		\For{NT, prods \textbf{in} grammar}
		\For{\textbf{each} production rule \textit{i} \textbf{of} prods}
		\If{$random() < $ prob\_mutation}
		\State value = $random\_Gaussian$(0, sd\_normal\_dist)
		\State $prob_i = min(prob_i $ + value, 1.0)
		\State $prob_i = max(prob_i $, 0.0)
		\State \textit{adjust\_probabilities}(prods)
		\State \textbf{break}
		\EndIf
		\EndFor
		\EndFor
		\EndProcedure
	\end{algorithmic}
\end{algorithm}

\begin{figure}[h!]
	\scalebox{0.7}{\fbox{%
			\parbox{0.1\textwidth}{%
				\begin{align*}
					{<}\text{expr}{>} ::=  & \, {<}\text{expr}{>}{<}\text{op}{>}{<}\text{expr}{>} & (0.73)\\
					& | \, {<}\text{var}{>} & \textbf{(0.27)}\\
					{<}\text{op}{>} ::= & \,+ \, & (0.25)\\
					& | \, - & (0.25)\\
					& | \, * & (0.25)\\
					& | \, / & (0.25)\\
					{<}\text{var}{>} ::= &  \, \text{x} & \textbf{(0.45)} \\ 
					& | \, \text{y} & (0.27)  \\
					& | \, \text{1.0} & (0.27)
				\end{align*}}}}
	\caption{PCFG grammar after mutation.}
	\label{mutcopgeafter}
\end{figure}

Fig. \ref{mutcopgeafter} shows an example of the grammar of Fig. \ref{copsgemapgram} after suffering a mutation. Assuming that the non-terminal $<expr>$ production rule $<var>$ has been randomly selected to be mutated, and that the random number generator of a Gaussian distribution of mean 0 and standard deviation 0.50 ($N(0, 0.50)$) has generated the number $-0.23$, the new probability of that production becomes 0.27 ($0.50-0.23$). The probability of the other production of the non-terminal $<expr>$ is adjusted to 0.73.
As a maximum of one mutation can occur in one production of each non-terminal, other mutations can occur in other non-terminals, for example, if the production rule $x$ of the non-terminal $<var>$ suffers a mutation of $+0.12$, it will have a value of 0.45, and the remaining outputs need to be adjusted, having a probability of 0.27. 


\subsection{Variation Operators}
\label{sec:cross}

Mutation and crossover are two genetic operators that can be used to change individuals.
The mutation operator modifies randomly chosen codons.
The codons are subjected to a Gaussian mutation, with the resultant value falling within the range [0,1].

\begin{figure}[h!]
	\centering
		\centering
\textbf{Genotype before mutation:}		\scalebox{0.7}{\framebox[3cm]{$<expr>$}\framebox[2cm]{$<op>$}\framebox[2.5cm]{$<var>$}} \\
		\scalebox{0.7}{\framebox[3cm]{[0.29, 0.73, 0.52]}\framebox[2cm]{[0.86]}\framebox[2.5cm]{[\textbf{0.41}, 0.15]}} \\

		\centering
		\textbf{\\Genotype after mutation:}
		\scalebox{0.7}{\framebox[3cm]{$<expr>$}\framebox[2cm]{$<op>$}\framebox[2.5cm]{$<var>$}} \\
		\scalebox{0.7}{\framebox[3cm]{[0.29, 0.73, 0.52]}\framebox[2cm]{[0.86]}\framebox[2.5cm]{[\textbf{0.64}, 0.15]}} \\
		
	\caption{Example of Co-PSGE's mutation on one codon of the genotype.}
	\label{fig:mutationpsge}
\end{figure}

Fig. \ref{fig:mutationpsge} shows an example of Gaussian mutation in an individual. Assuming that randomly only the first codon in the list of the non-terminal $<$var$>$ was selected for mutation and that the value generated with a Gaussian distribution of mean 0 and standard deviation 0.50 ($N(0,0.50)$) was $0.23$, the codon will be changed to $0.64$ ($0.41+0.23$).

The crossover operator is the same as the \gls{sge}'s and uses two parents to generate the offspring. A binary mask of the size of the number of lists of the genotype (the same number of non-terminals of the grammar) is randomly generated and the genotype of the offspring is created according to the values of the mask.

Fig. \ref{fig:xoversge} shows an example of a crossover between two individuals, generating one offspring and the randomly generated mask for the procedure. In the example shown, the descendant inherits the non-terminal $<$expr$>$ list from the second parent, and the remaining ones from the first. In case two descendants are generated, the other would get the opposite lists.

\begin{figure}[htbp]
	\centering
		\centering
		\textbf{Parents genotype:}\\
		Parent 1:\\
		\scalebox{0.7}{\framebox[3cm]{$<expr>$}\framebox[2cm]{$<op>$}\framebox[2.5cm]{$<var>$}} \\
		\scalebox{0.7}{\framebox[3cm]{[0.29, 0.73, 0.52]}\framebox[2cm]{[0.86]}\framebox[2.5cm]{[0.41, 0.15]}} \\
		\textbf{ }\\
		Parent 2:\\
		\scalebox{0.7}{\framebox[3cm]{$<expr>$}\framebox[2cm]{$<op>$}\framebox[3cm]{$<var>$}} \\
		\scalebox{0.7}{\framebox[3cm]{[0.16,0.71,0.48]}\framebox[2cm]{[0.23]}\framebox[3cm]{[0.19,0.86,0.56]}} \\
		
		\centering
		\textbf{\\Mask and offspring after crossover:}\\
		Mask:\\
		\scalebox{0.7}{\framebox[2cm]{$<expr>$}\framebox[2cm]{$<op>$}\framebox[2cm]{$<var>$}} \\
		\scalebox{0.7}{\framebox[2cm]{0}\framebox[2cm]{1}\framebox[2cm]{0}} \\
		\textbf{ }\\
		Offspring:\\
		\scalebox{0.7}{\framebox[3cm]{$<expr>$}\framebox[2cm]{$<op>$}\framebox[2.5cm]{$<var>$}} \\
		\scalebox{0.7}{\framebox[3cm]{[0.29, 0.73, 0.52]}\framebox[2cm]{[0.23]}\framebox[2.5cm]{[0.41, 0.15]}} \\

	\caption{Example of Co-PSGE's crossover between two individuals, generating one offspring.}
	\label{fig:xoversge}
\end{figure}

%% file: text/performance.tex
\section{Experimental Setup}
\label{sec:set}

The performance of \gls{copsge} will be compared with the results of \gls{ge} \cite{ONeill2003,Ryan1998,handbookge}, \gls{pge} \cite{Megane2021} and \gls{sge} \cite{Lourenco2016,Loureno2018}, analyzing the evolution of the mean best fitness in 100 independent runs. The framework of Whigham et al. \cite{Whigham2015} is followed, and four problems of different scopes are considered taking into account the recommendations of McDermott et al. \cite{McDermott2012}.
Table \ref{table:parameters} presents the parameters used by all the approaches.

Regarding the \gls{ge} and \gls{pge} variation operators, one-point crossover is used. The \gls{ge} mutation replaces the selected codons with new ones randomly generated on the interval [0, 255] and in the case of \gls{pge} a float mutation is used in which the codons are replaced with new ones generated on the interval [0, 1]. Neither of these methods uses a wrap mechanism.
Regarding the operators used by \gls{sge} and \gls{copsge}, these methods use the same crossover, which is the \gls{sge} crossover \cite{Loureno2018}. The mutation operator, in the case of \gls{sge}, replaces the selected codon with another valid option, that is, the index of another production of the same non-terminal, and in the case of \gls{copsge} a Gaussian mutation is performed with N(0, 0.5) on the value of the selected codon, keeping the new value in the range [0, 1].

Additionally, in \gls{pge} a learning factor of 1\% was used, and in \gls{copsge} a 5\% probability of a mutation occur in each non-terminal of the grammar, with a random value drawn from a $N(0, 0.50)$.

\begin{table}[h!]
\centering
\caption{\label{table:parameters}Parameters used in the experimental analysis for GE, PGE, SGE and Co-PSGE.}
\resizebox{0.7\columnwidth}{!}{%

\begin{tabular}{ccccc}
\multicolumn{1}{l}{} & GE             & PGE             & SGE        & Co-PSGE      \\
Population Size      & \multicolumn{4}{c}{{1000}}            \\
Generations          & \multicolumn{4}{c}{{50}}              \\
Elitism Count        & \multicolumn{4}{c}{{100}}             \\
Mutation Rate        & \multicolumn{4}{c}{{0.05}}            \\
Crossover Rate       & \multicolumn{4}{c}{{0.90}}            \\
Tournament           & \multicolumn{4}{c}{{3}}               \\
Size of Genotype     & \multicolumn{2}{c}{{128}} & \multicolumn{2}{c}{-}  \\
Max Depth            & \multicolumn{2}{c}{-}            & \multicolumn{2}{c}{10}
\end{tabular}
}
\end{table}

The fitness functions used to evaluate the individuals were designed with the objective of minimizing the error. In the case of Symbolic Regression and classification problems, the fitness is the \gls{rrse} between the individual's solution and the target on a data set. For the Boolean functions, the error is the number of incorrect predictions, and for the Path finding problem, the fitness is the number of pieces remaining after exceeding the step limit. All the problems are detailed bellow as well as the grammars used.


\subsection{Symbolic Regression}
\label{g:symb}
Popular benchmark problem for testing \gls{gp} algorithms, with the objective of finding the mathematical expression that best fits a given dataset. The Pagie polynomial was selected as it is a challenging symbolic regression problem \cite{pagie}, that has the following mathematical expression:

\begin{equation}
	\frac{1}{1+x[1]^{-4}}+\frac{1}{1+x[2]^{-4}} .
\end{equation}

The function is sampled with x[1], x[2] $\in$ [-5, 5] with a step of 0.4. The solutions are generated using the following grammar:

\begin{grammar}    
    \small
	
	<start> ::= <expr>
	
	<expr> ::= <expr> <op> <expr> | ( <expr> <op> <expr> )
	\alt <pre\_op> ( <expr> ) | <var>
	
	<op> ::= + | - | * | /
	
	<pre\_op> ::= sin | cos | exp | log | inv
	
	<var> ::= x[..] | 1.0
\end{grammar}

where $inv = \frac{1}{f(x)}$. The division and logarithm functions are protected, i.e., $1/0 = 1$ and $log(f(x)) = 0$ $ if f(x) \le 0$.

\subsection{Boston Housing}
This is a famous Machine Learning problem to predict the prices of Boston Houses. The dataset comes from the StatLib Library \cite{BostonHousing} and has 506 entries, with 13 features. It was divided in 90\% for training and 10\% for testing. The grammar used for the Boston Housing regression problem is the same as the Symbolic Regression (Section \ref{g:symb}).

	
	
	
	
	

\subsection{5-bit Even Parity}
The objective of this problem is to evolve a boolean function that takes as input a binary string with length 5, and returns 0 if the string is even or 1 if it is odd. Considering $b0$, $b1$, $b2$, $b3$, and $b4$ the input bits, the following grammar is used:

\begin{grammar}    
	    \small

	<start> ::= <B>
	
	<B> ::= <B> and <B> | <B> or <B> 
	\alt not (<B> and <B>) | not (<B> or <B>)
	\alt <var>
	
	<var> ::= b0 | b1 | b2 | b3 | b4 
\end{grammar}

\subsection{11-bit Boolean Multiplexer}
The aim of the 11-bit Multiplexer is to decode a 3-bit binary address and return the value of the corresponding data register ($d0$ to $d7$). The function receives as input three addresses ($s0$ to $s2$) and eight data registers ($i0$ to $i7$). For this problem we used:

\begin{grammar}    
	    \small

	<start> ::= <B>
	
	<B> ::= <B> and <B> | <B> or <B> | not <B>
	\alt <B> if <B> else <B> | <var>
	
	<var> ::= s0 | s1 | s2 | i0 | i1 | i3 | i4| i5 | i6 | i7
\end{grammar}

%% file: text/results.tex
\section{Results}
\label{sec:res}
To be able to support our study and compare the different approaches, a statistical analysis was conducted. Since the populations were independently initialized and the results do not meet the criteria for the parametric tests, the Kruskal-Wallis non-parametric test was employed to check for meaningful differences between the different methods. When this happened the Mann-Whitney post-hoc test with Bonferroni correction was performed to verify in which pairs the differences exist, with a level of significance of $\alpha = 0.05$. Table \ref{statstable} shows the statistical results, and values in bold indicate that \gls{copsge} outperforms the other approaches with statistical differences.

\begin{table}[h!]
\centering
\caption{\label{statstable}Results of the Mann-Whitney post-hoc statistical tests. The Bonferroni correction is used considering a significance level of $\alpha$ = 0.05. Values in bold mean that Co-PSGE is statistically better than GE, PGE or SGE.}
\resizebox{0.7\columnwidth}{!}{%
    \begin{tabular}{cccc}
        Problems              & GE     & PGE    & SGE    \\
        Pagie Polynomial     & \textbf{0.003} & 0.023          & 0.155          \\
        Boston Housing Train & \textbf{0.000} & \textbf{0.000} & 0.223          \\
        Boston Housing Test  & \textbf{0.000} & \textbf{0.045} & \textbf{0.000} \\
        5-bit Even Parity         & \textbf{0.000} & \textbf{0.000} & \textbf{0.000} \\
        11-bit Multiplexer       & \textbf{0.000} & \textbf{0.000} & 0.000  
    \end{tabular}}
\end{table}

Additionally, the effect size $r$ was calculated in order to determine how significant the differences are, and shown in Table \ref{effectsize}. The following notation was used: "\texttildelow" was used when there are no significant differences between samples, the "+" sign was used when the effect size is small ($r <= 0.3$), "++" was used when the effect size is medium ($0.3 < r <= 0.5$), and "+++" was used when the effect size is large ($r > 0.5$).

\begin{table}[h!]
\centering
\caption{\label{effectsize}Effect size between Co-PSGE and GE, PGE and SGE.}
\resizebox{0.65\columnwidth}{!}{%
    \begin{tabular}{cccc}
         Problems     & GE     & PGE    & SGE   \\
        Pagie Polynomial     &  +  &  -  &  \texttildelow          \\
        Boston Housing Train & +++ &  ++ &  \texttildelow          \\
        Boston Housing Test  & +++ &  +  &  ++         \\
        5-bit Even Parity         & +++ & +++ &  +++        \\
        11-bit Multiplexer       & +++ & +++ &  - - -      
    \end{tabular}}
\end{table}



\begin{figure}[h!]
	\centering
	\includegraphics[height=5cm]{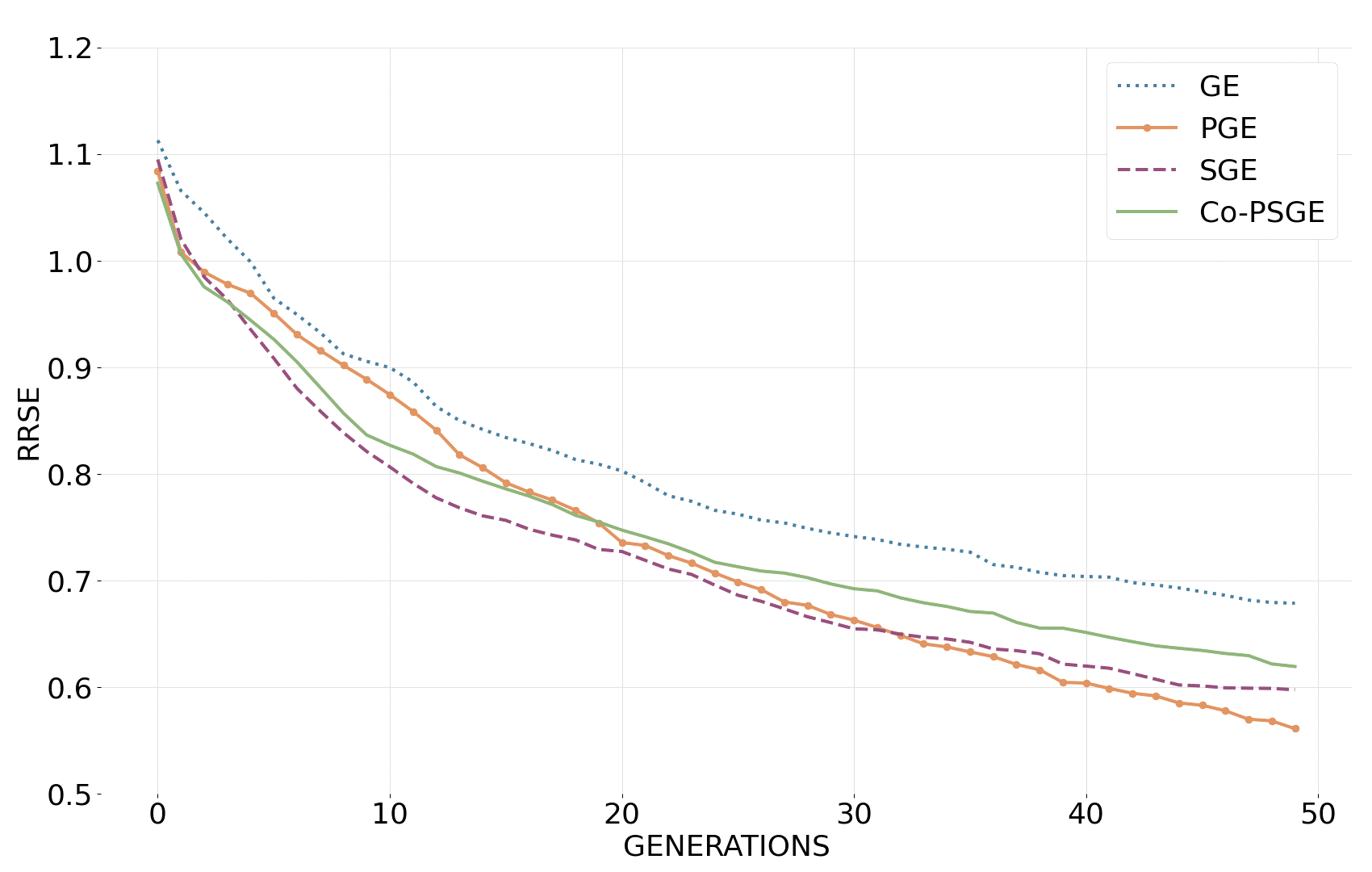}
	\caption{Performance results for the Pagie polynomial. Results are the mean best fitness of 100 runs.}
	\label{fig:pagieplot}
\end{figure}

The evolution of the mean best fitness over the 50 generations for the symbolic regression problem is represented in Fig. \ref{fig:pagieplot}. 
In this method \gls{pge} stands out by finishing with better fitness. These results are in line with what was observed in the work of Mégane et al. \cite{Megane2021}, in which \gls{pge} presented no statistical differences when compared with \gls{sge}. We see that although they all start with approximately the same fitness, \gls{pge} takes a few generations to keep up with the faster decrease of \gls{copsge} and \gls{sge}, however, it outperforms them after 20 and 30 generations, respectively. \gls{copsge} is significantly different from \gls{pge} on this problem, with a small effect size.
Comparing the performance of \gls{copsge} with that of \gls{ge}, we see that \gls{copsge} shows better results, presenting statistically significant differences with a large effect size. Compared to \gls{sge} we see that there are no significant differences between the methods.

\begin{figure}[h!]
	\centering
	\includegraphics[height=5cm]{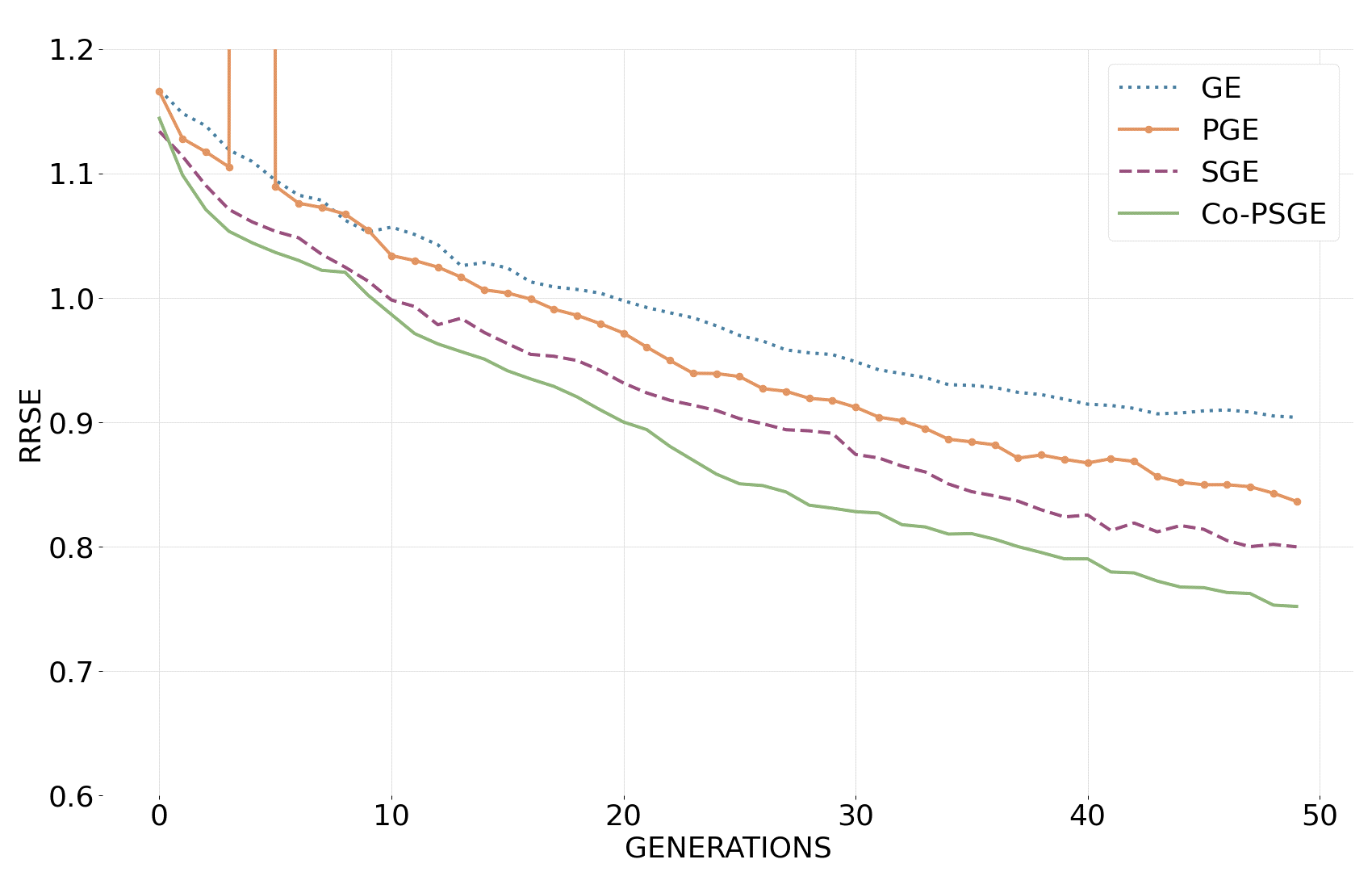}
	\caption{Testing results for the Boston Boston Housing problem. Results are the mean best fitness of 100 runs.}
	\label{fig:bhplot}
\end{figure}

The results in Fig. \ref{fig:bhplot} show the evolution of mean best fitness for the Boston Housing test data. For this problem we can see that the best performing method is \gls{copsge}, having a steeper decrease than the other approaches.
Looking at the statistical test results in Table \ref{statstable} and effect size in Table \ref{effectsize}, we can see that \gls{copsge} is statistically better than all methods, showing a large effect size relative to \gls{ge}, a small effect size relative to \gls{pge}, and a medium effect size relative to \gls{sge}.
The training data is interesting to analyze, since in training \gls{sge} and \gls{copsge} do not show significant differences, however in testing they do exist with a medium effect size, which shows that \gls{copsge} creates a more generalized model for the unseen data.


\begin{figure}[h!]
	\centering
	\includegraphics[height=5cm]{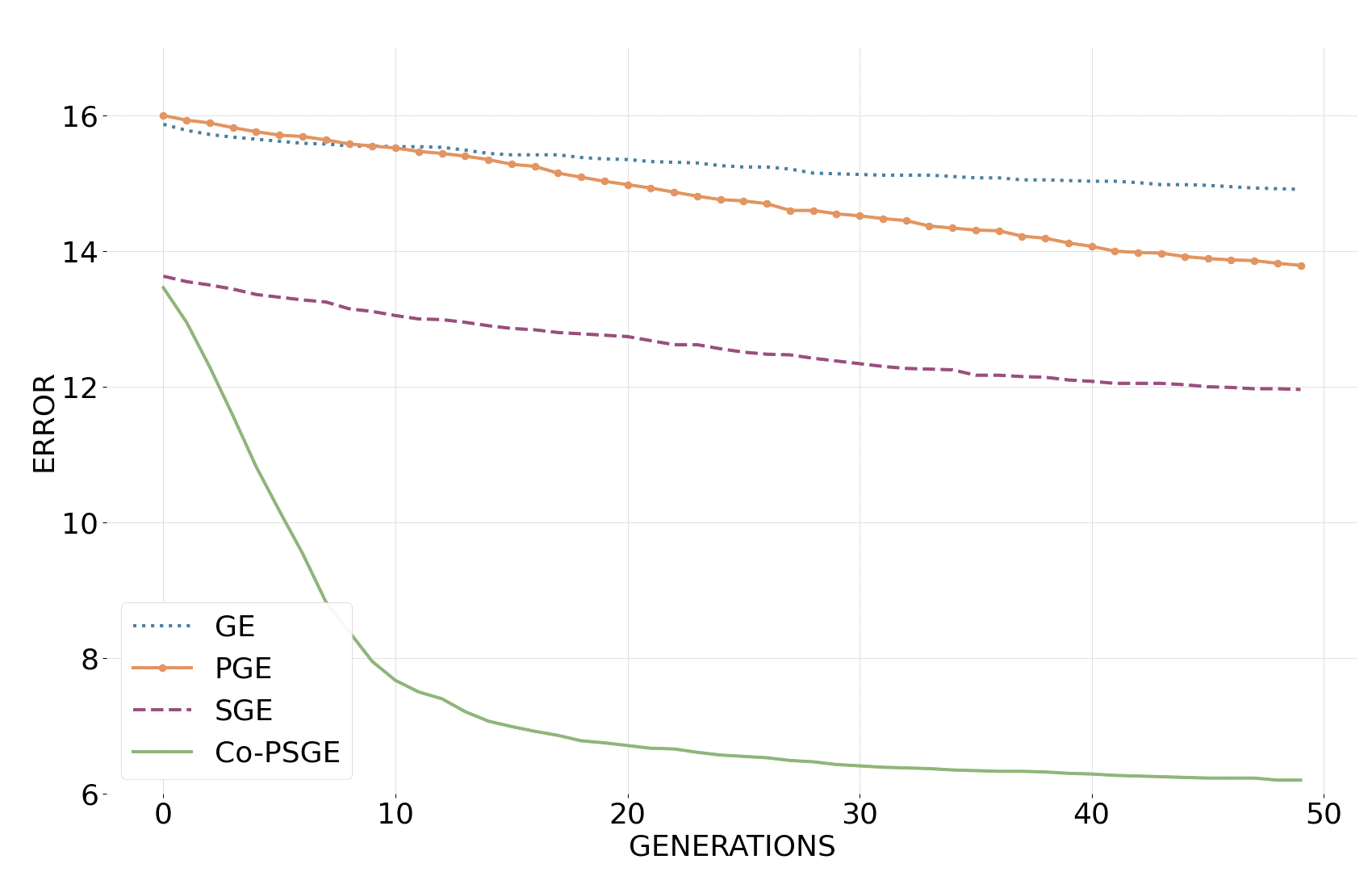}
	\caption{Performance results for the 5-Bit Even Parity problem. Results are the mean best fitness of 100 runs.}
	\label{fig:5parplot}
\end{figure}

Looking at the results in Fig. \ref{fig:5parplot} for the 5-bit parity problem, we observe that \gls{sge} and \gls{copsge} start the evolutionary process with better average fitness when compared to \gls{ge} and \gls{pge}. However, the decrease slope of \gls{copsge} is much steeper, distancing itself quickly from the curve of \gls{sge}, and in a few generations it manages to reach a better error more than twice as much as the other approaches. \gls{copsge} shows significant differences from all methods, with a large effect size.
Analyzing the performance of \gls{ge} and \gls{pge}, we see that \gls{pge} has a more pronounced decrease, however in the analyzed generations it does not reach the behavior of \gls{sge}.
An analysis was performed on the average probability of the grammars of the best individuals in each generation of all the runs, and we observed that the probability of the NOR production rule ends with probability greater than 90\%.
Since this is a universal gate, through it we can reach the remaining productions and consequently the best solution.


\begin{figure}[h!]
	\centering
	\includegraphics[height=5cm]{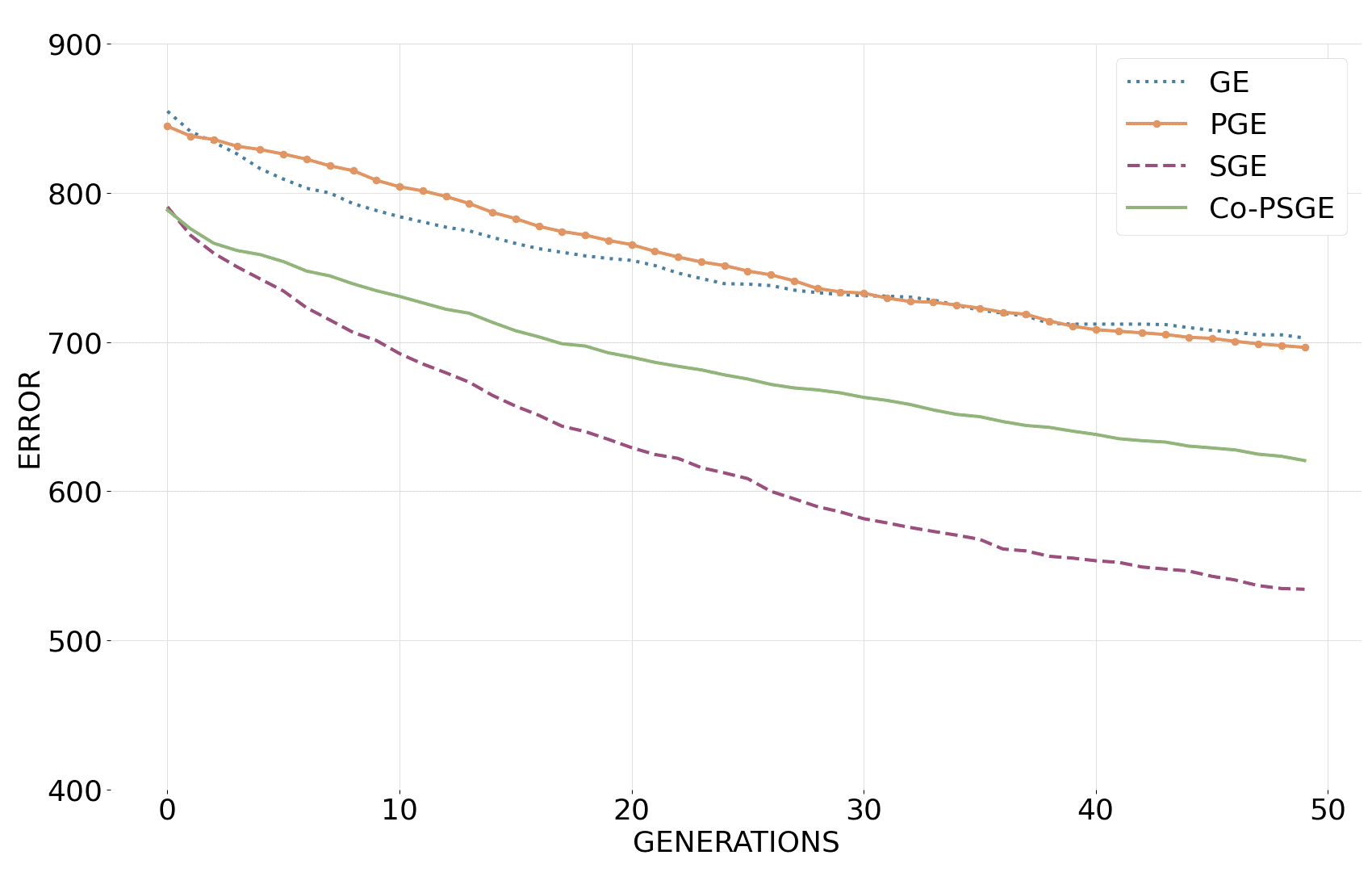}
	\caption{Performance results for the 11-bit Boolean Multiplexer problem. Results are the mean best fitness of 100 runs.}
	\label{fig:11multplot}
\end{figure}

The results for the 11-bit multiplexer are shown in Fig. \ref{fig:11multplot}. We see that \gls{ge} and \gls{pge} start with worse fitness compared to \gls{sge} and \gls{copsge}, and maintain throughout the evolutionary process, with \gls{copsge} showing statistically significant differences with a large effect size for both methods. 
Although they start with the same fitness, \gls{sge} has a larger decrement, which makes it statistically different on this problem compared to \gls{copsge}.





%% file: text/conc.tex
\section{Conclusion}
\label{sec:conc}
In this paper we proposed \gls{copsge}, a method that adapts the representation and mapping of \gls{sge} by assigning each individual a grammar that, like the individual, is subjected to variation operators. The genotype is a list of dynamic lists, each of which corresponds to a non-terminal of the grammar and contains real numbers, which correspond to the probability of choosing a derivation rule.

The proposed method was compared with standard \gls{ge}, \gls{pge} and \gls{sge} on four different benchmark problems. Statistical tests showed that \gls{copsge} is better than \gls{ge}, outperforming it with significant differences on all problems. \gls{pge} was outperformed in three of the four problems analyzed, while \gls{sge} was outperformed in two problems.
The co-evolution of the grammar with the genotype of the individuals allowed guiding the evolutionary process to better solutions faster, as shown by the performance analysis, supporting the work presented by O'Neill et al. \cite{ONeillRyan2004}. Further experimentation will be necessary to gain more insight into the advantages of the proposed method.

As future work it will be interesting to study different metrics, such as locality and redundancy of \gls{copsge}, in order to be able to analyze the impact that mutated grammars have on the representation of individuals and compare it with the results of \gls{ge} and \gls{sge}.
Another line of work will be to analyze the evolution of the grammars by initializing them with different probabilities for each individual, but also test different grammars, following the work of Nicolau et al. \cite{Nicolau2018}. It would also be interesting to test problems from different scopes, including multi-optimization problems.